\documentclass[journal]{IEEEtran}
\usepackage{enumitem}
\usepackage[colorlinks, citecolor=blue, urlcolor=blue, bookmarks=false, hypertexnames=true]{hyperref}
\usepackage{amsmath}
\usepackage{amssymb}
\usepackage{xcolor}
\usepackage{booktabs} 
\usepackage{colortbl}
\usepackage{multirow} 
\usepackage{graphicx}
\usepackage{subfigure}
\usepackage{makecell}

\begin{document}
\title{AFFormer: Adaptive Feature Fusion Transformer for V2X Cooperative Perception under\\ Channel Impairments}

\author{
    {Xi Zhou,
    Tao Huang\IEEEauthorrefmark{1},~\IEEEmembership{Senior Member,~IEEE},
    Qing-Long Han,~\IEEEmembership{Fellow,~IEEE},
    Rana Abbas, \\
    Mostafa Rahimi Azghadi,~\IEEEmembership{Senior Member,~IEEE}
    }
    
\thanks{\IEEEauthorrefmark{1}Corresponding author.}

\thanks{X.~Zhou, T.~Huang, and M. R. Azghadi are with the College of Science and Engineering, James Cook University, Smithfield, QLD 4878, Australia. (e-mail: xi.zhou@jcu.edu.au, tao.huang1@jcu.edu.au, mostafa.rahimiazghadi@jcu.edu.au.}

\thanks{Q.-L.~Han is with the School of Engineering, Swinburne University of Technology, Melbourne, VIC 3122, Australia (e-mail: qhan@swin.edu.au).}

\thanks{R. Abbas is with Transport for NSW, Sydney, NSW, Australia. (e-mail: Rana.Abbas@transport.nsw.gov.au).}

\thanks{The code is available at \url{https://github.com/zoeyzhouxi/AFFormer}.}

\thanks{This work has been submitted to the IEEE for possible publication. Copyright may be transferred without notice, after which this version may no longer be accessible.}
}

% The paper headers
\markboth{ }
{\MakeLowercase{\textit{et al.}}: Demo of IEEEtran.cls for IEEE Journals}

\maketitle

\begin{abstract}

Accurate 3D object detection is essential for ensuring the safety of autonomous vehicles. 
Cooperative perception, which leverages vehicle-to-everything (V2X) communication to share perceptual data, enhances detection but is vulnerable to channel impairments, such as noise, fading, and interference. 
To strengthen the reliability of intelligent transportation systems, this work improves the robustness of V2X cooperative perception under communication conditions that reflect common channel impairments.
This paper proposes an Adaptive Feature Fusion Transformer (AFFormer), a Transformer-based framework that mitigates the adverse effects of corrupted features by modeling temporal, inter-agent, and spatial correlations. 
AFFormer introduces three key modules: Multi-Agent and Temporal Aggregation for context-aware fusion across agents and over time, Dual Spatial Attention for efficient modeling of spatial dependencies, and Uncertainty-Guided Fusion for entropy-driven refinement of fused features. 
A teacher–student knowledge distillation strategy further enhances robustness by aligning fused features with reliable early-collaboration supervision. 
\textcolor{black}{
AFFormer is validated on the V2XSet and DAIR-V2X datasets, where it consistently outperforms existing methods under both ideal and impaired communication conditions, demonstrating improved robustness to communication-induced feature degradation while maintaining a competitive efficiency-accuracy trade-off.}

\end{abstract}

\begin{IEEEkeywords}
Vehicle-to-everything,
cooperative perception,
artificial intelligence,
multi-agent systems,
connected and autonomous vehicles,
autonomous driving
\end{IEEEkeywords}

\section{Introduction}
Reliable detection of surrounding road participants constitutes the foundation for subsequent motion planning and vehicle control tasks, significantly influencing the overall driving safety of intelligent vehicles \cite{ref1}. 
With the rapid advancement of large-scale datasets and powerful computer vision algorithms, single-vehicle perception systems, in which autonomous vehicles independently perceive their environment using onboard sensors, have achieved considerable progress \cite{ref2}. 
However, these single-vehicle systems inherently face limitations, particularly in scenarios involving occluded or distant objects. 
Such limitations hinder reliable decision-making in safety-critical situations, underscoring the need for approaches that extend perception capabilities beyond those of individual vehicles.
To address this issue, vehicle-to-everything (V2X) cooperative perception (CP) enables multiple agents, including autonomous vehicles and roadside infrastructure, to share complementary perceptual data, thereby extending the perceptual range, improving detection accuracy, and providing a more comprehensive and reliable understanding of the surrounding environment \cite{ref3}.
Such enhanced situational awareness is essential for the safety and reliability of intelligent transportation systems.

Information sharing in V2X CP can be categorized into early, late, and intermediate collaboration, distinguished by the extent of data processing before transmission 
\cite{ref3}.
Early collaboration exchanges raw sensor data among agents, providing rich information but demanding substantial bandwidth and strict synchronization, which is difficult under limited or unstable network conditions \cite{ref4}.
Late collaboration shares only final perception outputs, such as object classifications and locations, thereby greatly reducing communication costs but limiting data granularity and adaptability in complex scenarios \cite{ref5}.
Intermediate collaboration transmits pre-processed features, striking a balance by reducing data volume while retaining essential information and offering an effective trade-off between detection accuracy and bandwidth requirements \cite{ref6}.
{\color{black}To meet bandwidth constraints, many V2X CP systems adopt compact intermediate features for transmission, making the subsequent fusion process highly sensitive to corrupted or degraded messages.

In practice, transmitted features are inevitably affected by channel impairments arising from the physical and networking characteristics of vehicular wireless links \cite{ref7}. 
Large-scale attenuation and shadowing arise from distance-dependent path loss and dynamic blockages, whereas multipath propagation and high mobility introduce small-scale fading and Doppler shifts, leading to rapid channel fluctuations and burst errors  \cite{ref8}.
Furthermore, interference and medium contention in shared spectrum environments increase packet error rates \cite{ref9}.
In real-time V2X cooperative perception, retransmission is often impractical due to strict latency requirements and rapidly changing traffic environments. 
Consequently, relying solely on perfectly received messages or discarding degraded shared information can significantly reduce the availability of useful cooperative context. 
Importantly, even corrupted feature tensors may still contain partially informative cues that can benefit perception.
However, such distortions are particularly detrimental in intermediate-feature collaboration, where even minor perturbations can propagate through the fusion pipeline, leading to feature misalignment, confidence miscalibration, and ultimately missed detections or false alarms. 
This challenge is not adequately addressed by most existing methods, which typically assume ideal communication conditions. 
As a result, detection consistency and reliability may be significantly compromised. 
Therefore, it is essential to develop robust fusion mechanisms that suppress unreliable components while preserving informative content, thereby maximizing the utility of shared features under imperfect communication conditions.}

To this end, we propose an Adaptive Feature Fusion Transformer (AFFormer), a fusion mechanism designed to operate under imperfect communication conditions by adaptively downweighting unreliable components while preserving informative cues from degraded shared features, thereby supporting more reliable CP in connected and autonomous vehicles.  
By explicitly modeling temporal, inter-agent, and spatial correlations, AFFormer is able to recover useful information from imperfectly received features and suppress noise-induced distortions, thereby enhancing resilience and improving perception accuracy under dynamic and challenging communication conditions.
We demonstrate the effectiveness of AFFormer by integrating it into a CP framework and benchmarking it against several state-of-the-art models on the V2XSet \cite{ref10} and DAIR-V2X \cite{ref11} datasets.  
The main contributions of this paper are summarized as follows:  
\begin{itemize}
\item A Multi-Agent and Temporal Aggregation (MATA) module is introduced to jointly capture inter-agent and temporal dependencies, producing unified fusion features. By assigning adaptive attention weights, MATA effectively suppresses irrelevant or corrupted information while aggregating complementary cues from neighboring agents and historical frames, leading to more context-aware representations.  
\item A Dual Spatial Attention (DualSA) module is developed to efficiently model spatial interactions by decomposing attention computation along two orthogonal dimensions (width and height). This dual-branch structure substantially reduces computational complexity compared with conventional self-attention while preserving rich spatial context.  
\item An Uncertainty-Guided Fusion (UGF) module is designed to adaptively refine spatial features obtained from DualSA. Leveraging entropy-based uncertainty estimation, UGF generates spatially varying importance maps that emphasize informative regions and suppress noisy or ambiguous patterns, producing a more robust and discriminative spatial representation.  
\textcolor{black}{
\item Extensive experiments on V2XSet and DAIR-V2X demonstrate that AFFormer consistently improves detection accuracy and robustness under communication impairments while maintaining strong performance under ideal communication conditions and a competitive efficiency-accuracy trade-off.}

\end{itemize}

The remainder of this paper is structured as follows. Section~\ref{related_work} reviews relevant literature on CP and communication constraints. Section~\ref{method} formalizes the problem setting, details the proposed fusion architecture, and describes the knowledge distillation training strategy. Section~\ref{experiment} presents the experimental setup, including datasets, implementation details, evaluation metrics, and baseline comparisons, followed by comprehensive experimental results and analysis. Finally, Section~\ref{conclusion} concludes the paper with a summary of key findings and contributions.

\section{Related Works}\label{related_work}

\subsection{V2X CP}
CP enabled by V2X communication offers a promising solution to overcome the limitations of single-vehicle perception.  
Based on the stage at which information is shared and integrated, CP can be categorized into early, intermediate, and late fusion, among which intermediate fusion has received significant attention for its balance between accuracy and efficiency.  

A variety of intermediate fusion methods have been proposed for application in intelligent transportation systems.  
For example, F-Cooper \cite{ref12} employs max pooling to fuse voxel-level features, while CoCa3D \cite{ref13} enhances camera-based detection by applying point-wise maximum fusion to BEV features from multiple connected autonomous vehicles \cite{ref14}.  
Multi-agent collaboration can also be represented as a graph, where nodes denote agent states and edges indicate pairwise interactions or shared information \cite{ref15}.  
CoAlign \cite{ref16} improves pose alignment among agents through a pose graph optimization framework using an Agent-Object Pose Graph to enhance spatial consistency.  
Similarly, a graph attention fusion network \cite{ref17} applies spatial and channel-wise attention mechanisms to emphasize critical regions during feature aggregation.

Transformer-based mechanisms have also been adopted to dynamically assign feature fusion weights.  
V2VFormer \cite{ref18} and V2VFormer++ \cite{ref19} leverage spatial and channel attention to integrate features across agents.  
MKD-Cooper \cite{ref20} applies multi-view knowledge distillation, transferring information from multiple teacher models into a unified student model.  
DSRC \cite{ref21} addresses feature corruption caused by weather, sensor failure, and interference using semantic-guided sparse-to-dense distillation and feature-to-point reconstruction.  
Where2comm \cite{ref22} employs multi-scale feature attention to integrate information from all agents.  

Despite these advancements, most methods assume ideal communication and overlook common impairments, such as noise, interference, and data corruption.  
Consequently, their performance deteriorates under practical communication constraints, limiting their applicability in deployed systems.  

\subsection{V2X CP under Communication Constraints}
{\color{black}In practical deployments, V2X CP must operate under various communication constraints, including limited bandwidth, transmission delays, and wireless channel impairments, all of which significantly affect the reliability, timeliness, and safety of CP systems.

Due to limited wireless bandwidth, transmitting raw sensor data among agents is often infeasible.
To reduce communication overhead while preserving essential semantic information, many studies compress intermediate features before transmission.}
For instance, DiscoNet~\cite{ref23} compresses features using a $1\times1$ convolutional autoencoder, and V2X-ViT~\cite{ref10} employs multiple $1\times1$ convolutions to reduce feature dimensionality.
COOPERNAUT~\cite{ref24} encodes point clouds with a Point Transformer-based encoder, while What2comm~\cite{ref25} filters redundant background features to prioritize informative regions.
However, these approaches adopt fixed compression ratios, limiting adaptability under dynamic network conditions.
To address this, SmartCooper~\cite{ref26} introduces a learnable encoder that adjusts compression based on channel state information and incorporates a gating mechanism to suppress detrimental inputs.

{\color{black}Beyond bandwidth constraints, CP is also susceptible to communication latency arising from network congestion, scheduling delays, and wireless transmission overhead.
Since received features may correspond to earlier timestamps rather than the current frame, temporal misalignment can introduce spatial inconsistencies during feature fusion and degrade the perception of dynamic objects.
Several methods explicitly compensate for this through temporal alignment or asynchronous fusion mechanisms. 
V2X-ViT~\cite{ref10} introduces delay-aware positional encoding to capture inter-agent temporal offsets, while V2VNet~\cite{ref27} incorporates relative pose and delay time during feature aggregation.
SyncNet~\cite{ref28} predicts current-frame features from historical observations using a dual-branch LSTM, and CoBEVFlow~\cite{ref29} estimates BEV flow maps to spatially align asynchronous messages.
V2X-PC~\cite{ref30} takes an alternative approach, bypassing historical features entirely and predicting object positions directly from clustered point representations.

While bandwidth and latency have received considerable attention, the impact of wireless channel impairments on transmitted feature representations remains relatively under-explored.
In practice, channel fading and noise can distort feature tensors even when packets are successfully decoded, yielding corrupted representations that degrade fusion performance.
Several recent works have begun to address this issue.  
V2VAM+LCRN~\cite{ref31} reconstructs corrupted features using pixel-specific convolution kernels, but it models degradation with uniformly distributed random noise, which oversimplifies realistic channel effects.
To better approximate practical communication environments, \cite{ref32} simulates Rician fading with free-space path loss and proposes a self-supervised adaptive weighting mechanism to mitigate channel distortion, though it does not explicitly model temporal dependencies.
V2X-INCOP~\cite{ref33} employs spatio-temporal convolutions to recover features disrupted by communication interruptions but assumes that historical features remain complete and uncorrupted.
RoCooper~\cite{ref34} further incorporates communication-aware modeling and leverages multi-dimensional feature correlations with an ego-feature anchoring mechanism for feature recovery and fusion, demonstrating improved robustness under impairment conditions.}

\section{Proposed Method} \label{method}
This study addresses cooperative LiDAR-based 3D object detection for autonomous driving, focusing on the practical challenge of feature corruption caused by communication channel impairments in V2X systems.
V2X CP is inherently complex and involves additional factors such as communication delays and localization errors.
To isolate the specific impact of transmission corruption, these factors are considered beyond the scope of this study.
To mitigate the adverse impact of corrupted data on cooperative detection, a novel Transformer-based adaptive feature fusion framework, termed AFFormer, is proposed.
This section first presents an overview of the V2X CP system, followed by a detailed description of the proposed fusion framework.
Finally, the teacher–student training strategy for enhancing feature recovery and model robustness is introduced.

\subsection{Problem Formulation of V2X CP}\label{PF}

\begin{figure}[!t]
    \centering
    \includegraphics[width=0.8\columnwidth]{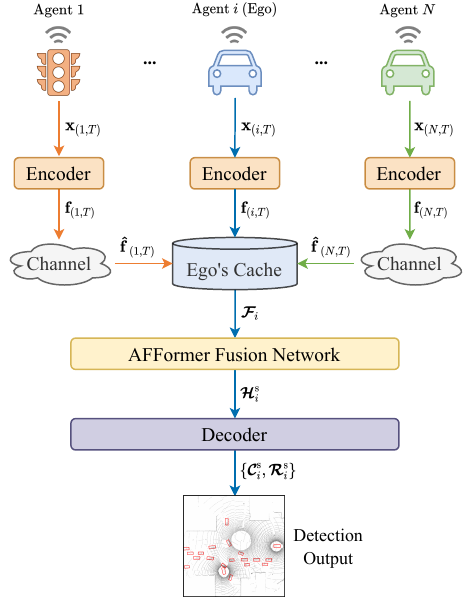}
    \caption{Overview of the proposed AFFormer-based V2X CP system. Each agent encodes its local LiDAR observations and transmits the resulting features to the ego vehicle via wireless communication channels. The ego vehicle aggregates its own and received features in a cache, after which AFFormer adaptively integrates inter-agent, temporal, and spatial information to produce fused representations for the detection head.}
    \label{fig:v2xcp}
\end{figure}

As illustrated in Fig.~\ref{fig:v2xcp}, we consider a V2X CP system comprising $N$ agents, each capable of precise pose estimation and maintaining a local memory buffer of $T$ feature frames. 
Without loss of generality, agent $i$ is designated as the ego agent.
At timestep $t$, the raw LiDAR observation collected by agent $i$ is denoted $\mathbf{x}_{(i,t)}$, which is first processed by a PointPillar encoder~$f_{\mathrm{encode}}$~\cite{ref35} to extract a high-level feature representation:
\begin{equation}
\mathbf{f}_{(i,t)} = f_{\mathrm{encode}}(\mathbf{x}_{(i,t)})\in \mathbb{R}^{H \times W \times C},
\end{equation}
where $H \times W$ denotes the spatial dimensions of the feature map and $C$ is the number of channels.

{\color{black}
At each timestep, agents transmit their encoded features to the ego vehicle via a wireless channel subject to impairments, which introduce distortions that cause the received feature to deviate from the transmitted one.
We simulate feature transmission using a communication-theoretic formulation that incorporates large-scale path loss, small-scale fading, and additive noise.
For large-scale attenuation, the free-space path loss (FSPL) model is used for vehicle-to-infrastructure (V2I) links, while the WINNER~II model is used for vehicle-to-vehicle (V2V) links to account for multipath effects and environmental variability.
Small-scale fading is modeled as Rician fading for V2I and highway scenarios, and as Rayleigh fading for urban V2V scenarios.

The received signal power at the ego vehicle from agent $j$ ($j \neq i$) at timestep $t$ is computed  via a standard link-budget model:
\begin{equation}
P_{\mathrm{rx},(j,t)}
=
P_{\mathrm{tx}}
+ G_{\mathrm{tx}}
+ G_{\mathrm{rx}}
- \mathrm{PL}\!\left(d_{(j,t)},\, f_\mathrm{c}\right),
\label{eq:rx_power}
\end{equation}
where $P_{\mathrm{tx}}$ is the transmit power, $G_{\mathrm{tx}}$ and $G_{\mathrm{rx}}$ are the transmit and receive antenna gains, $\mathrm{PL}(\cdot)$ is the large-scale path loss in dB, $d_{(j,t)}$ is the inter-agent distance, and $f_\mathrm{c}$ is the carrier frequency.
The receiver noise power over bandwidth $B$ is
\begin{equation}
P_{\mathrm{n}}
= N_0 + 10\log_{10}(B) + \mathrm{NF},
\label{eq:noise_power}
\end{equation}
where $N_0$ is the thermal noise power spectral density and $\mathrm{NF}$ is the receiver noise figure in dB.
The received SNR is then
\begin{equation}
\mathrm{SNR}_{(j,t)}
= P_{\mathrm{rx},(j,t)} - P_{\mathrm{n}}.
\label{eq:snr_db}
\end{equation}

In our feature-domain channel simulation, to avoid unrealistically shrinking feature magnitudes by absolute path-loss scaling, we apply a \emph{relative} large-scale attenuation normalized by a reference distance $d_0$:
\begin{equation}
G_{\mathrm{rel}}\!\left(d_{(j,t)}\right)
\triangleq
10^{-\frac{\mathrm{PL}(d_{(j,t)})-\mathrm{PL}(d_0)}{20}}.
\label{eq:greldef}
\end{equation}
The noise-free received feature is then
\begin{equation}
\mathbf{s}_{(j,t)}
\triangleq
G_{\mathrm{rel}}\!\left(d_{(j,t)}\right)\,
|h_{(j,t)}|\,\mathbf{f}_{(j,t)},
\end{equation}
where $h_{(j,t)}$ is the complex small-scale fading coefficient.
The received feature power $P_{s,(j,t)}$, estimated by the empirical mean over feature elements during the simulation, can be described as
\begin{equation}
P_{s,(j,t)}
\triangleq
\mathbb{E}\!\left[\left\|\mathbf{s}_{(j,t)}\right\|_2^2\right].
\end{equation}
We inject element-wise AWGN $\mathbf{n}_{(j,t)}\sim \mathcal{N}(0,\sigma^2_{(j,t)})$ and set the variance to match the target SNR as
\begin{equation}
\sigma^2_{(j,t)}
= \frac{P_{s,(j,t)}}{10^{\mathrm{SNR}_{(j,t)}^{\mathrm{dB}}/10}}.
\label{eq:sigma_from_snr}
\end{equation}
The received (potentially corrupted) feature can be formulated as
\begin{equation}
\hat{\mathbf{f}}_{(j,t)}
= \mathbf{s}_{(j,t)} + \mathbf{n}_{(j,t)}.
\end{equation}

The cooperative feature tensor at timestep $t$, aggregated by the ego agent, is
\begin{equation}
\mathbf{F}_{(i,t)}
= \bigl[\hat{\mathbf{f}}_{(1,t)},\,\ldots,\,
         \mathbf{f}_{(i,t)},\,\ldots,\,
         \hat{\mathbf{f}}_{(N,t)}\bigr]^{\!\mathrm{T}}
\in \mathbb{R}^{N \times H \times W \times C},
\end{equation}
where the ego's own feature $\mathbf{f}_{(i,t)}$ is generated locally and is not subject to wireless channel impairments.}
Across consecutive timesteps, cooperative features form a feature sequence:  
\begin{equation}
\boldsymbol{\mathcal{F}}_i = [\mathbf{F}_{(i,1)}, \mathbf{F}_{(i,2)}, \ldots, \mathbf{F}_{(i,T)}] \in \mathbb{R}^{N \times T \times H \times W \times C}.
\end{equation}

This sequence serves as the input to AFFormer, denoted as \(f_\mathrm{fuse}\), which generates a unified and robust representation:  
\begin{equation}
\boldsymbol{\mathcal{H}}^{\mathrm{s}}_i = f_\mathrm{fuse}\left(\boldsymbol{\mathcal{F}}_i \right) \in \mathbb{R}^{H \times W \times C}.
\end{equation}  
The primary objective of AFFormer is to suppress the detrimental effects of corrupted features in CP.  
Architectural details are presented in Section~\ref{aff-net}.  

Following feature fusion, the representation is processed by the decoding module \(f_{\mathrm{decode}}\), consisting of two convolutional branches: one for foreground–background classification and the other for bounding-box regression.  
The detection outputs are expressed as:  
\begin{equation}
\{\smash[b]{\boldsymbol{\mathcal{C}}}^{\mathrm{s}}_i, \smash[b]{\boldsymbol{\mathcal{R}}}^{\mathrm{s}}_i\} = f_{\mathrm{decode}}(\boldsymbol{\mathcal{H}}^{\mathrm{s}}_i),
\end{equation}  
where \(\smash[b]{\boldsymbol{\mathcal{C}}}^{\mathrm{s}}_i\) denotes classification predictions (object likelihoods), and \(\smash[b]{\boldsymbol{\mathcal{R}}}^{\mathrm{s}}_i\) denotes regression predictions for 3D bounding-box parameters (position, dimensions, and yaw angle).  

\subsection{Adaptive Feature Fusion Transformer (AFFormer)}
\label{aff-net}

\begin{figure*}[th]
\centering
\includegraphics[width=\textwidth]{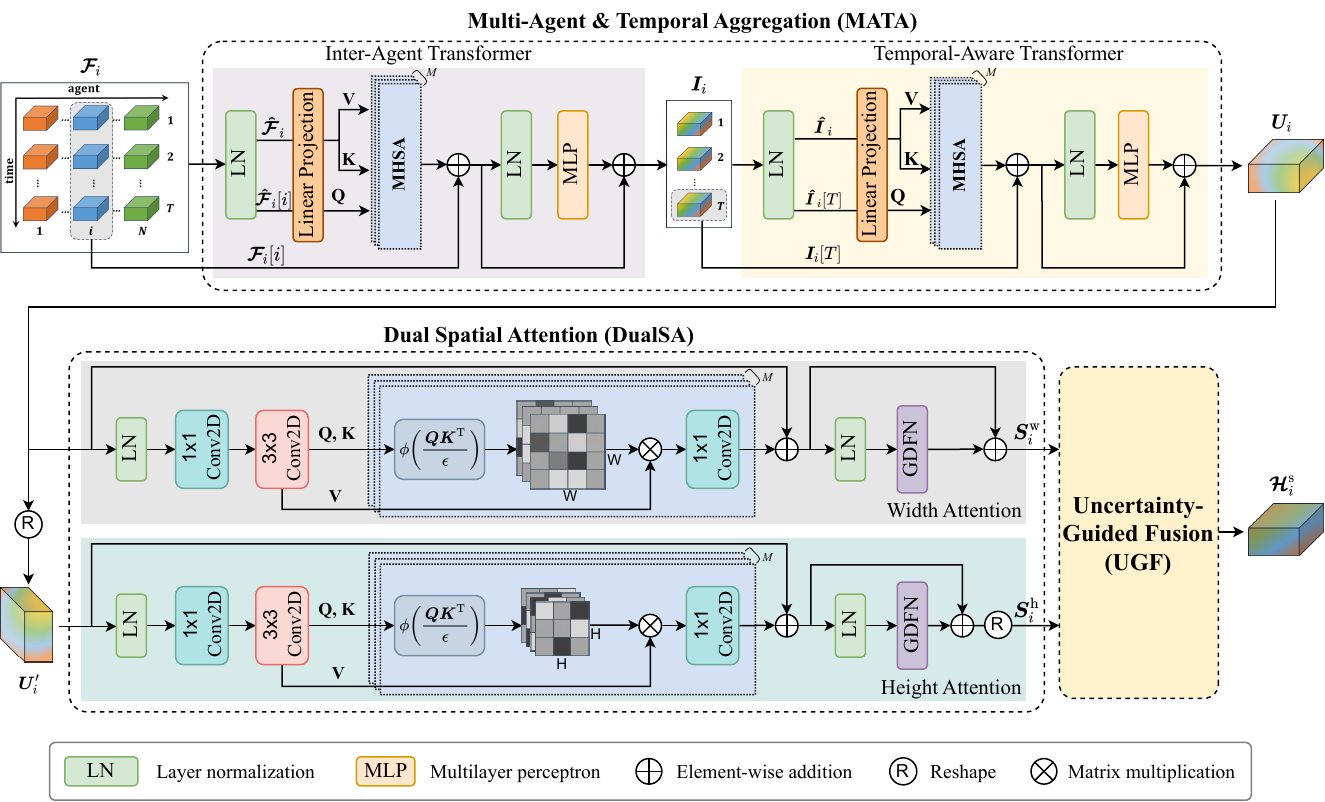}
\caption{Overall architecture of AFFormer, comprising three core modules.
(1)~\textit{Multi-Agent and Temporal Aggregation (MATA)}: integrates 
inter-agent and temporal information via cascaded Inter-Agent Transformer 
(IAT) and Temporal-Aware Transformer (TAT) blocks to produce a unified 
intermediate representation.
(2)~\textit{Dual Spatial Attention (DualSA)}: models spatial dependencies 
along the width and height dimensions independently, reducing computational 
complexity relative to global self-attention while preserving fine-grained 
spatial context.
(3)~\textit{Uncertainty-Guided Fusion (UGF)}: adaptively combines 
spatial-aware features using entropy-based uncertainty weighting to suppress 
corrupted or unreliable information and produce robust fused representations.}
\label{fig:afformer}
\end{figure*}

{\color{black}
AFFormer addresses feature fusion from three complementary perspectives: inter-agent interaction, temporal consistency, and spatial dependency.
\textit{Inter-agent fusion} aggregates feature representations received from multiple collaborating agents via V2X communication.
\textit{Temporal fusion} exploits feature correlations across consecutive timesteps to improve perception stability under dynamic conditions.
\textit{Dual spatial attention} models spatial dependencies along two orthogonal directions, enhancing spatial representation while remaining computationally efficient.
\textit{Uncertainty-guided fusion} adaptively weights features according to their estimated reliability, suppressing degraded information during aggregation.

As illustrated in Fig.~\ref{fig:afformer}, these four perspectives are realized through three principal modules: MATA, DualSA, and UGF, which jointly mitigate feature corruption by exploiting inter-agent, temporal, and spatial correlations.
The MATA module processes the input feature sequence through two cascaded Transformer blocks, an Inter-Agent Transformer (IAT) and a Temporal-Aware Transformer (TAT), to incorporate contextual cues from neighboring agents and historical frames, producing a unified intermediate representation 
$\boldsymbol{U}_i$.
The DualSA module then enhances spatial context by independently capturing dependencies along the width and height dimensions via two parallel Transformer branches, generating spatially attentive features $\boldsymbol{S}_{i}^{\mathrm{w}}$ and $\boldsymbol{S}_{i}^{\mathrm{h}}$; this decomposed design reduces complexity from $\mathcal{O}(H^2W^2)$ to $\mathcal{O}(HW(H+W))$ relative to global self-attention.
Finally, the UGF module adaptively integrates $\boldsymbol{S}_{i}^{\mathrm{w}}$ and $\boldsymbol{S}_{i}^{\mathrm{h}}$ by assessing pixel-wise feature reliability through entropy-based uncertainty estimation, refining informative regions while suppressing corrupted patterns to yield the robust fused representation $\boldsymbol{\mathcal{H}}_i^{\mathrm{s}}$.
The detailed design of each module is presented in the following subsections.}

\subsubsection{Multi-Agent and Temporal Aggregation (MATA) Module}\label{mata_s}

The MATA module is specifically designed to mitigate feature corruption by exploiting spatial and temporal correlations within the CP framework.  
It integrates complementary information from spatially distributed agents and leverages temporal continuity across historical frames, thereby improving the robustness and accuracy of the ego agent's feature representation.  
To achieve this, we employ two cascaded Transformer blocks: the IAT, which aggregates spatially correlated features from neighboring agents, and the TAT, which models dependencies across sequential timesteps.  
Both IAT and TAT share a common architectural backbone based on multi-head self-attention (MHSA), enabling consistent and efficient feature fusion across agents and over time.   

For brevity, the MHSA-based feature fusion process is abstracted as a general function \(f_{\mathrm{MHSA}}(\cdot)\).  
Given the projected queries, keys, and values \(\{\boldsymbol{Q}_m, \boldsymbol{K}_m, \boldsymbol{V}_m\}_{m=1}^{M}\) for all \(M\) heads and a residual reference feature \(\boldsymbol{R}\) (e.g., the ego feature or current feature), the fused feature \(\boldsymbol{F}\) is defined as:  
\begin{equation}
    \boldsymbol{F}=f_{\mathrm{MHSA}}\left ( \{\boldsymbol{Q}_m, \boldsymbol{K}_m, \boldsymbol{V}_m\}_{m=1}^{M}, \boldsymbol{R}\right ),
\label{eqmhsa}
\end{equation}
where \(f_{\mathrm{MHSA}}(\cdot)\) includes scaled dot-product attention across all heads, concatenation, output projection, residual connection, and feed-forward refinement.

The MHSA process internally performs the following steps.  
For each head \(m\), scaled dot-product attention is first applied:  
\begin{equation}
    \boldsymbol{H}_m = \phi \left( \frac{\boldsymbol{{Q}}_m {\boldsymbol{{K}}_{m}^{\mathrm{T}}}}{\sqrt{d_k}} \right) \boldsymbol{{V}}_m, \label{mhsaeq1}
\end{equation}
where \(\phi(\cdot)\) is the softmax function and \(d_k\) denotes the dimensionality of each head.  
Outputs from all heads are concatenated along the channel dimension and projected back to the original dimensionality \(C\): 
\begin{equation}
     \boldsymbol{\breve{F}}=[\boldsymbol{H}_1;\boldsymbol{H}_2;\dots;\boldsymbol{H}_M]\boldsymbol{W},
\end{equation}
where \([;]\) denotes channel-wise concatenation and \(\boldsymbol{W}\in \mathbb{R}^{M\cdot d_k \times C}\) is a learnable projection matrix.  
A residual connection is then applied by adding the squeezed \(\boldsymbol{\breve{F}}\) to the reference feature \(\boldsymbol{R}\): 
\begin{equation}
    \boldsymbol{\hat{F}}=\mathrm{Squeeze}(\boldsymbol{\breve{F}}) + \boldsymbol{R},
\end{equation}
where \(\mathrm{Squeeze}(\cdot)\) removes singleton dimensions.  
Finally, refinement is performed via a feed-forward network \(f_{\mathrm{MLP}}(\cdot)\) following layer normalization \(\mathrm{LN}(\cdot)\):  
\begin{equation}
    \boldsymbol{F}= \boldsymbol{\hat{F}} + f_{\mathrm{MLP}} \left ( \mathrm{LN}(\boldsymbol{\hat{F}})\right ).
\end{equation}

As shown at the top of Fig.~\ref{fig:afformer}, the IAT is first applied within the MATA module.    
Formally, the original ego feature \(\boldsymbol{{\mathcal{F}}}_i[i] \in \mathbb{R}^{T\times D \times C}\), with spatial dimension \(D=H\times W\), serves as the reference feature.  
Layer normalization is applied to the cooperative feature sequence \(\boldsymbol{\mathcal{F}}_i\) to obtain \(\boldsymbol{\hat{\mathcal{F}}}_i=\mathrm{LN}(\boldsymbol{\mathcal{F}}_i)\in \mathbb{R}^{N\times T \times D \times C}\).  
The normalized ego feature \(\boldsymbol{\hat{\mathcal{F}}}_i[i]\) is used as the query, while the full normalized sequence \(\boldsymbol{\hat{\mathcal{F}}}_i\) provides keys and values.  
Linear transformations then generate the projections for each head:  %
\begin{align}
&\boldsymbol{Q}_m^{\mathrm{ia}} = \boldsymbol{\hat{\mathcal{F}}}_i[i] \boldsymbol{W}_m^\mathrm{Q} \in \mathbb{R}^{T\times D \times d_k} \overset{\text{reshaping}}{\rightarrow} \mathbb{R}^{T \times D \times 1\times d_k}, \\
&\boldsymbol{K}_m^{\mathrm{ia}} = \boldsymbol{\hat{\mathcal{F}}}_i \boldsymbol{W} _m^\mathrm{K} \in \mathbb{R}^{N\times T \times D \times d_k} \overset{\text{reshaping}}{\rightarrow} \mathbb{R}^{T \times D \times N\times d_k}, \\
&\boldsymbol{V}_m^{\mathrm{ia}} = \boldsymbol{\hat{\mathcal{F}}}_i \boldsymbol{W} _m^\mathrm{V} \in \mathbb{R}^{N\times T \times D \times d_k} \overset{\text{reshaping}}{\rightarrow} \mathbb{R}^{T \times D \times N \times d_k},
\end{align}
where \(\boldsymbol{W}_m^\mathrm{Q}, \boldsymbol{W}_m^\mathrm{K}, \boldsymbol{W}_m^\mathrm{V} \in \mathbb{R}^{C \times d_k}\) are learnable parameters shared across attention components, and reshaping aligns dimensions for matrix multiplication.  
Following Eq.~(\ref{eqmhsa}), the inter-agent fusion feature \(\boldsymbol{I}_i \in \mathbb{R}^{T \times D \times C}\) is obtained as:  
\begin{equation}
    \boldsymbol{I}_i=f_{\mathrm{MHSA}}\left ( \{\boldsymbol{Q}_m^{\mathrm{ia}}, \boldsymbol{K}_m^{\mathrm{ia}}, \boldsymbol{V}_m^{\mathrm{ia}}\}_{m=1}^{M}, \boldsymbol{{\mathcal{F}}}_i[i]\right ).
\end{equation}

The resulting \(\boldsymbol{I}_i\) is subsequently processed by the TAT, which refines the feature by exploiting temporal correlations across frames.  
Here, the current inter-agent fusion feature \(\boldsymbol{I}_i[T]\in \mathbb{R}^{D \times C}\) serves as the reference.  
The input \(\boldsymbol{I}_i\) is normalized as \(\boldsymbol{\hat{I}}_i=\mathrm{LN}(\boldsymbol{I}_i)\), from which linear projections generate queries, keys, and values for each head:  
\begin{align}
& \boldsymbol{Q}_m^{\mathrm{ta}} = \boldsymbol{\hat{I}}_i[T]\boldsymbol{\hat{W}} _m^\mathrm{Q} \in \mathbb{R}^{D \times d_k}  \overset{reshaping}{\rightarrow} \mathbb{R}^{D \times 1\times d_k} ,   \\
& \boldsymbol{K}_m^{\mathrm{ta}} = \boldsymbol{\hat{I}}_i \boldsymbol{\hat{W}} _m^\mathrm{K}\in \mathbb{R}^{T \times D \times d_k} \overset{reshaping}{\rightarrow} \mathbb{R}^{D \times T\times d_k},  \\
& \boldsymbol{V}_m^{\mathrm{ta}} = \boldsymbol{\hat{I}}_i \boldsymbol{\hat{W}} _m^\mathrm{V}\in \mathbb{R}^{T \times D \times d_k} \overset{reshaping}{\rightarrow} \mathbb{R}^{D \times T \times d_k},
\end{align} 
where \(\boldsymbol{\hat{W}} _m^\mathrm{Q}, \boldsymbol{\hat{W}} _m^\mathrm{K}, \boldsymbol{\hat{W}} _m^\mathrm{V}\in \mathbb{R}^{C\times d_k}\) are learnable parameters.  
Finally, following the MHSA fusion process, the unified inter-agent and temporal feature \(\boldsymbol{U}_i \in \mathbb{R}^{D \times C}\) is obtained:  
\begin{equation}
    \boldsymbol{U}_i=f_{\mathrm{MHSA}}\left ( \{\boldsymbol{Q}_m^{\mathrm{ta}}, \boldsymbol{K}_m^{\mathrm{ta}}, \boldsymbol{V}_m^{\mathrm{ta}}\}_{m=1}^{M}, \boldsymbol{I}_i[T]\right ).
\end{equation}

\subsubsection{Dual Spatial Attention (DualSA) Module}
The DualSA module captures pixel interactions in \(\boldsymbol{U}_i\) across spatial dimensions to enhance representation learning and mitigate the effects of corrupted information.  
Unlike conventional spatial attention mechanisms, which explicitly compute pairwise pixel interactions, DualSA decomposes attention computation into two orthogonal branches, width and height attention, as illustrated at the bottom of Fig.~\ref{fig:afformer}.  
In the Width Attention branch, the unified feature \(\boldsymbol{U}_i \in \mathbb{R}^{H \times W \times C}\) is first normalized, \(\boldsymbol{\hat{U}}_i=\mathrm{LN}(\boldsymbol{U}_i)\), and then transformed via a point-wise \(1\times 1\) convolution followed by a depthwise \(3\times 3\) convolution.  
The output is evenly split along the channel dimension into query, key, and value components:  
\begin{equation} \label{a1}
    [\boldsymbol{Q}, \boldsymbol{K}, \boldsymbol{V}]=\mathrm{Chunk}\left(\mathrm{Conv}_{3\times 3}(\mathrm{Conv}_{1\times 1}(\boldsymbol{\hat{U}}_i))\right),
\end{equation}
where \(\mathrm{Chunk}(\cdot)\) denotes a channel-wise partitioning operation, yielding three tensors of dimension \(\mathbb{R}^{H \times W \times C'}\). 
Here, \(C' = M \times C_\mathrm{head}\), where \(M\) is the number of attention heads, and \(C_\mathrm{head}\) is the feature dimension per head.

\begin{figure*}[th]
\centering
\includegraphics[width=0.9\textwidth]{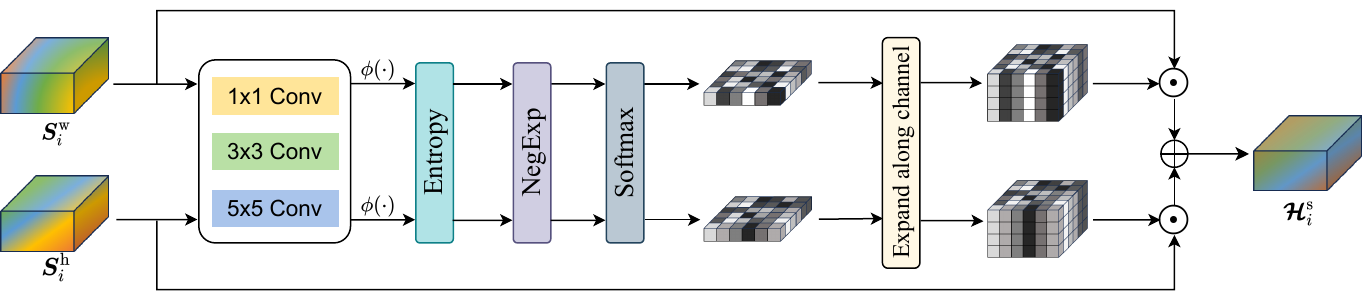} 
\caption{Architecture of the \textit{Uncertainty-Guided Fusion (UGF)} module. Width- and height-attended features are first processed through multi-scale convolutional layers to extract local spatial context. Entropy-based uncertainty estimation is then applied, followed by negative exponential and softmax operations to generate spatial importance weights. These weights are expanded along the channel dimension and used to adaptively emphasize reliable regions and suppress uncertain areas, resulting in robust fused features.}
\label{fig:ugf}
\end{figure*}

Query, key, and value are reshaped into \(\boldsymbol{\hat{Q}}, \boldsymbol{\hat{K}}, \boldsymbol{\hat{V}} \in \mathbb{R}^{M\times H \times W \times C_\mathrm{head}}\) for multi-head attention.  
The width-aware embedding is obtained through scaled dot-product attention, followed by a reshaping operation that concatenates the outputs from all attention heads:  
\begin{equation}
    \boldsymbol{E}_i = \phi \left ( \frac{\boldsymbol{\hat{Q}}\boldsymbol{\hat{K}}^{\mathrm{T}}}{\epsilon} \right ) \boldsymbol{\hat{V}} \in \mathbb{R}^{M\times H \times W \times C_\mathrm{head}} \overset{reshaping}{\rightarrow} \mathbb{R}^{H \times W \times C'},
\end{equation}
where \(\epsilon\) is a learnable scaling parameter. Then, the width-aware embedding is projected back to the original dimension through a \(1\times 1\) convolution and added to the input \(\boldsymbol{U}_i\) for residual stability: 
\begin{equation}
    \boldsymbol{\hat{E}}_i = \mathrm{Conv}_{1\times 1}(\boldsymbol{{E}}_i) + \boldsymbol{U}_i.
\end{equation}
Finally, \(\boldsymbol{\hat{E}}_i\) is normalized and enhanced through the Gated Depthwise Convolutional Feed-forward Network (GDFN) \cite{ref36}, which suppresses uninformative responses while refining salient features.  
The width-aware spatial representation is, therefore, given by:  
\begin{equation} \label{a5}
    \boldsymbol{{S}}_{i}^\mathrm{w} = f_\mathrm{GDFN}(\mathrm{LN}(\boldsymbol{\hat{E}}_i)) + \boldsymbol{\hat{E}}_i.
\end{equation}

The Height Attention branch follows an analogous procedure.  
The unified feature \(\boldsymbol{U}_i \in \mathbb{R}^{H \times W \times C}\) is first transposed to \(\boldsymbol{U'}_i\in \mathbb{R}^{W \times H \times C}\), allowing attention computation along the height axis.  
The resulting height-aware feature \(\smash[b]{\boldsymbol{S'}}_{i}^{\mathrm{h}} \in \mathbb{R}^{H \times W \times C}\) is obtained by the same processing steps and subsequently rearranged back to the original spatial configuration, yielding \(\boldsymbol{S}_{i}^{\mathrm{h}} \in \mathbb{R}^{H \times W \times C}\).

By decomposing 2D attention into width and height branches executed in parallel, DualSA effectively captures anisotropic spatial dependencies while avoiding the high computational cost of full 2D self-attention.  
The integration of multi-head attention and GDFN-based refinement in each branch further strengthens feature discrimination and robustness.  
The outputs \(\boldsymbol{S}_{i}^{\mathrm{w}}\) and \(\boldsymbol{S}_{i}^{\mathrm{h}}\) are subsequently combined by the UGF module (Section~\ref{sec:ugf}), which adaptively weighs them according to spatial confidence.  
This yields a refined spatial representation that is both uncertainty-adaptive and robust, thereby providing a strong foundation for downstream CP tasks.

\subsubsection{Uncertainty-Guided Fusion (UGF) Module} \label{sec:ugf}

While DualSA effectively captures spatial dependencies through width and height attention, these two perspectives may contribute unequally depending on scene structure and feature confidence.  
Moreover, communication impairments may still introduce out-of-distribution noise, leading to residual uncertainty in object detection.
To address this, we propose the UGF module (Fig.~\ref{fig:ugf}), which adaptively fuses \(\boldsymbol{S}_{i}^{\mathrm{w}}\) and \(\boldsymbol{S}_{i}^{\mathrm{h}}\) based on their estimated spatial uncertainty.  
Instead of static or learnable scalar weights, UGF generates entropy-based, spatially varying importance maps that prioritize reliable regions while suppressing ambiguous or noisy patterns, thereby producing a refined and robust representation \(\boldsymbol{\mathcal{H}}_i^{\mathrm{s}}\).  

{\color{black}Let $\boldsymbol{S}_{i}^n \in \mathbb{R}^{H \times W \times C}$ denote the spatial-aware feature from branch $n \in \{\mathrm{w}, \mathrm{h}\}$.
To capture contextual information across multiple receptive fields, three parallel convolutional branches with kernel sizes $1\times1$, $3\times3$, and $5\times5$ are applied to $\boldsymbol{S}_{i}^n$, extracting fine-grained, mid-range, and large-context spatial patterns, respectively.
Each branch produces feature maps with identical spatial resolution and channel dimension $C$, ensuring proper alignment across scales.
The outputs are then aggregated via element-wise summation to form the multi-scale representation:
\begin{equation}
\hat{\boldsymbol{S}}_{i}^n =
\mathrm{Conv}_{1\times1}(\boldsymbol{S}_{i}^n)
+ \mathrm{Conv}_{3\times3}(\boldsymbol{S}_{i}^n)
+ \mathrm{Conv}_{5\times5}(\boldsymbol{S}_{i}^n).
\end{equation}

Channel responses at each spatial location encode distinct activation patterns captured by the convolutional filters.
Applying a softmax function $\phi(\cdot)$ across the channel dimension normalizes these responses into a categorical probability distribution that reflects the relative activation strength of each channel.
A learnable temperature parameter $\tau$ controls the sharpness of this distribution:
\begin{equation}
\boldsymbol{P}_{i}^n = \phi\!\left(\frac{\hat{\boldsymbol{S}}_{i}^n}{\tau}\right).
\end{equation}
If activations are concentrated on a small number of channels, the distribution is sharp, indicating high confidence in the encoded representation.
Conversely, a flatter distribution over many channels indicates greater ambiguity.
The Shannon entropy of this channel-wise distribution is used to quantify uncertainty at each spatial location:
\begin{equation}
\boldsymbol{\mathcal{E}}_i^n =
-\sum_{c=1}^{C}
\boldsymbol{P}_{i}^n[:,:,c]
\cdot \log\boldsymbol{P}_{i}^n[:,:,c],
\end{equation}
where higher entropy indicates greater uncertainty, which is subsequently used to guide uncertainty-aware feature fusion in the UGF module.}

The entropy map $\boldsymbol{\mathcal{E}}_{i}^n \in \mathbb{R}^{H \times W \times 1}$ 
is first normalised to $[0,1]$ via min--max scaling for numerical 
stability, and then converted to an importance map through an exponential 
decay function:
\begin{equation}
\boldsymbol{\mathcal{W}}_{i}^n =
\exp\!\left(
-\frac{
  \boldsymbol{\mathcal{E}}_{i}^n - \min(\boldsymbol{\mathcal{E}}_{i}^n)
}{
  \max(\boldsymbol{\mathcal{E}}_{i}^n) - \min(\boldsymbol{\mathcal{E}}_{i}^n)
}
\right),
\end{equation}
such that spatially certain regions (low entropy) receive high weights while uncertain or corrupted regions are down-weighted.
Each importance map is then expanded along the channel dimension to $\hat{\boldsymbol{\mathcal{W}}}_{i}^n \in \mathbb{R}^{H \times W \times C}$, and normalized across the two branches via softmax to yield relative importance scores:
\begin{equation}
\boldsymbol{\xi}_{i}^n =
\frac{
  \exp(\hat{\boldsymbol{\mathcal{W}}}_{i}^n)
}{
  \exp(\hat{\boldsymbol{\mathcal{W}}}_{i}^{\mathrm{w}})
  + \exp(\hat{\boldsymbol{\mathcal{W}}}_{i}^{\mathrm{h}})
}.
\end{equation}
The final fused representation is obtained by uncertainty-weighted 
summation:
\begin{equation}
\boldsymbol{\mathcal{H}}_i^{\mathrm{s}} =
\boldsymbol{\xi}_{i}^{\mathrm{w}} \odot \boldsymbol{S}_{i}^{\mathrm{w}}
+ \boldsymbol{\xi}_{i}^{\mathrm{h}} \odot \boldsymbol{S}_{i}^{\mathrm{h}},
\end{equation}
where $\odot$ denotes element-wise multiplication.

\subsection{System Training with Knowledge Distillation}

In V2X scenarios, channel impairments corrupt features received from neighboring agents, precluding reliable ground-truth supervision for these degraded representations.
To address this, we adopt a teacher-student knowledge distillation strategy~\cite{ref37}, in which an early-collaboration model serves as the teacher and the proposed AFFormer as the student.
The teacher provides clean features and detection pseudo-labels to guide student training; only the student is deployed at inference.

The teacher is built on the PointPillar backbone~\cite{ref35} and operates under ideal communication conditions.
At the current timestep $T$, raw point clouds from all $N$ agents, $\{\mathbf{x}_{(1,T)}, \ldots, \mathbf{x}_{(N,T)}\}$, are transformed into the ego coordinate frame and fused into a global point cloud $\mathbf{X}_{(i,T)}$. 
Following the same encoding and decoding pipeline as the student model, \(\mathbf{X}_{(i, T)}\) is processed to generate high-level features and detection outputs:  
\begin{align}
    &\boldsymbol{\mathcal{H}}^{\mathrm{t}}_i= f_{\mathrm{encode}}(\mathbf{X}_{(i,T)}) \in \mathbb{R}^{H\times W\times C},\\
    &\{\smash[b]{\boldsymbol{\mathcal{C}}}^{\mathrm{t}}_i, \smash[b]{\boldsymbol{\mathcal{R}}}^{\mathrm{t}}_i\}=f_{\mathrm{decode}}(\boldsymbol{\mathcal{H}}^{\mathrm{t}}_i).
\end{align}

The teacher is trained with a binary cross-entropy loss \(L_{\mathrm{ce}}\) for foreground--background classification and a smooth-$L_1$ loss \(L_{\mathrm{smooth}}\) for bounding-box 
regression:  
\begin{equation}
\mathcal{L}^{\mathrm{t}}= \sum_{i=1}^{N} \left ( \alpha L_{\mathrm{ce}}(\boldsymbol{\mathcal{C}}_i, \boldsymbol{\mathcal{C}}_i^t)+\beta L_{\mathrm{smooth}}(\boldsymbol{\mathcal{R}}_i, \boldsymbol{\mathcal{R}}_i^t) \right ),
\end{equation}
where $\boldsymbol{\mathcal{C}}_i$ and $\boldsymbol{\mathcal{R}}_i$ denote the ground-truth classification and regression targets within the perception region of agent $i$, and $\alpha$, $\beta$ are loss weighting coefficients.
\textcolor{black}{The teacher network is employed solely during training to provide a high-quality supervisory signal based on features obtained under ideal communication conditions, serving as an upper-bound reference for feature fusion.
This follows a common knowledge distillation paradigm, in which cleaner or less-corrupted representations are used to guide the learning of a more robust student model.
Since the teacher is used solely for training-time supervision and is discarded thereafter, AFFormer incurs no additional communication overhead and does not depend on ideal channel conditions during deployment. During inference, agents exchange only intermediate feature representations rather than raw point clouds, which maintains compatibility with practical V2X bandwidth constraints.}

The student model is trained with a joint objective that combines detection loss and knowledge distillation loss.  
The teacher’s features act as pseudo-labels to guide AFFormer’s fused features, enabling the reconstruction of corrupted inputs and improving robustness.  
Additional detection-level distillation further enhances accuracy.  
We adopt the Kullback–Leibler divergence \cite{ref38} \(L_{\mathrm{kl}}\) to quantify discrepancies between teacher and student outputs.  
The knowledge distillation loss is expressed as:  
\begin{equation}
\mathcal{L}_{\mathrm{kd}} = \sum_{i=1}^{N} \left ( 
L_{\mathrm{kl}}(\boldsymbol{\mathcal{H}}^{\mathrm{t}}_i, \boldsymbol{\mathcal{H}}^{\mathrm{s}}_i) 
+ L_{\mathrm{kl}}(\boldsymbol{\mathcal{C}}^{\mathrm{t}}_i, \boldsymbol{\mathcal{C}}^{\mathrm{s}}_i) 
+ L_{\mathrm{kl}}(\boldsymbol{\mathcal{R}}^{\mathrm{t}}_i, \boldsymbol{\mathcal{R}}^{\mathrm{s}}_i) 
\right ),
\end{equation}
where \(\boldsymbol{\mathcal{H}}^{\mathrm{s}}_i\), \(\boldsymbol{\mathcal{C}}^{\mathrm{s}}_i\), and \(\boldsymbol{\mathcal{R}}^{\mathrm{s}}_i\) denote the fused features and detection outputs of the student.

The detection loss \(\mathcal{L}_{\mathrm{det}}\) for the student mirrors that of the teacher and supervises each agent’s detection head with its own ground truth:  
\begin{equation}
    \mathcal{L}_{\mathrm{det}}=\sum_{i=1}^{N} \left ( \alpha L_{\mathrm{ce}}(\boldsymbol{\mathcal{C}}_i, \boldsymbol{\mathcal{C}}_i^\mathrm{s})+\beta L_{\mathrm{smooth}}(\boldsymbol{\mathcal{R}}_i, \boldsymbol{\mathcal{R}}_i^\mathrm{s}) \right ).
\end{equation}

Finally, the overall student loss is a weighted combination of detection and distillation objectives:  
\begin{equation}
    \mathcal{L}^{\mathrm{s} }= \gamma \mathcal{L}_{\mathrm{kd} }+\mathcal{L}_{\mathrm{det} },
\end{equation}
where \(\gamma\) is a hyperparameter that balances the influence of distillation. 

\section{Experiments} \label{experiment}

This section evaluates the effectiveness of the proposed method in LiDAR-based 3D object detection. 

\subsection{Datasets}

We conduct experiments on two publicly available CP datasets, covering both simulated and real-world scenarios.  

\subsubsection{V2XSet} 

V2XSet \cite{ref10} is an open-source V2X perception dataset generated via co-simulation with CARLA \cite{ref39} and OpenCDA \cite{ref40}.  
It comprises 55 scenarios across five roadway types and eight CARLA towns, with each scenario involving between two and seven intelligent agents.  
Each agent is equipped with a 32-channel LiDAR sensor with a 120-meter sensing range.  
The dataset contains 11,447 frames, split into 6,694 training, 1,920 validation, and 2,833 testing frames.  
To enable temporal dependency modeling, two historical frames are attached to each sample; only samples with at least two valid historical frames are retained.    

\subsubsection{DAIR-V2X}
DAIR-V2X~\cite{ref11} is a real-world V2X CP dataset collected in operational driving environments, providing accurate 3D bounding-box annotations.
Each scene comprises one ego vehicle and one roadside infrastructure unit, both equipped with a LiDAR sensor and a front-facing camera.
We use the cooperative subset DAIR-V2X-C, which contains 9{,}000 synchronized LiDAR frame pairs captured from vehicle and infrastructure viewpoints across 100 representative scenes.
\textcolor{black}{Following the standard cooperative perception protocol, the autonomous vehicle is designated as the ego agent, while the infrastructure unit serves as an auxiliary perception source providing complementary observations.
All agent features, including temporal information, are transformed into the ego-vehicle coordinate frame prior to fusion, ensuring consistent multi-agent temporal modeling within the MATA module.}
Only samples with at least one historical frame are retained to support learning of temporal dependencies.

\begin{table}[!t]
\renewcommand{\arraystretch}{1.3}
\caption{\textcolor{black}{Communication parameters used in the channel simulation.}}
\label{tab:comm_params}
\centering
\renewcommand\tabcolsep{4pt}
\begin{tabular*}{\columnwidth}{@{\extracolsep{\fill}}lcc}
% \begin{tabular}{lcc}
\toprule
\textbf{Parameters} & Notation & \textbf{Value}  \\
\midrule
Carrier frequency & $f_{\mathrm{c}}$ & 5.9 GHz \\
Bandwidth & $B$ & 10 MHz  \\
Transmit power & $P_{\mathrm{tx}}$ & 23 dBm \\
Vehicle transmitter antenna gain & $G_{\mathrm{tx}}^{\mathrm{v}}$ & 3 dBi  \\
RSU transmitter antenna gain & $G_{\mathrm{tx}}^{\mathrm{r}}$ & 6 dBi  \\
Receiver antenna gain & $G_{\mathrm{rx}}$ & 3 dBi  \\
Receiver hardware noise figure & $\mathrm{NF}$ & 6 dB  \\
Thermal noise density & $N_0$ & $-174$ dBm/Hz  \\
\bottomrule
\end{tabular*}
\end{table}

\begin{table*}[!t]
\renewcommand{\arraystretch}{1.15}
\caption{\textcolor{black}{Comparison of object detection performance under ideal and impaired communication conditions.}}
\label{tab:performance}
\centering
\renewcommand\tabcolsep{4pt}
\begin{tabular*}{\textwidth}{@{\extracolsep{\fill}}lccccccccc}
\toprule
\multirow{2.5}{*}{Method} & \multirow{2.5}{*}{\makecell[c]{Corruption\\aware}}
& \multicolumn{4}{c}{V2XSet (Ideal / Impaired)} 
& \multicolumn{4}{c}{DAIR-V2X (Ideal / Impaired )} \\
\cmidrule(l{1pt}r{2pt}){3-6}\cmidrule(l{2pt}r{1pt}){7-10}
&
& AP@0.5$\uparrow$ & $\Delta$AP@0.5$\downarrow$
& AP@0.7$\uparrow$ & $\Delta$AP@0.7$\downarrow$
& AP@0.5$\uparrow$ & $\Delta$AP@0.5$\downarrow$ 
&AP@0.7$\uparrow$ & $\Delta$AP@0.7$\downarrow$\\
\midrule
V2X-ViT \cite{ref10} & $\times$
& 91.21 / 87.83 & 3.38 & 83.07 / 80.55 
& 2.52
& 70.95 / 62.58 & 8.37 & 53.80 / 51.19
& 2.61 \\
CoAlign \cite{ref16} & $\times$
& 92.67 / 68.98 & 23.69 & 85.01 / 58.97
& 26.04
& 78.14 / 67.90 & 10.24 & 64.33 / 56.17
& 8.16 \\
MKD-Cooper \cite{ref20} & $\times$
& 91.38 / 85.73 & 5.65 & 82.08 / 78.13
& 3.95
& 77.38 / 68.29 & 9.09 & 63.69 / 55.31
& 8.38 \\
DSRC \cite{ref21} & $\surd$
& 92.43 / 70.82 & 21.61 & 86.31 / 55.03
& 31.28
& 75.34 / 68.15 & 7.19 & 63.16 / 54.06
& 9.10 \\
V2VAM+LCRN \cite{ref31} & $\surd$
& 94.93 / 75.62 & 19.31 & 85.43 / 59.88
& 25.55
& 70.84 / 66.71 & 4.13 & 54.46 / 52.71
& \textbf{1.75} \\
AFFormer (Ours) & $\surd$
& \textbf{96.44} / \textbf{93.34} & \textbf{3.1} & \textbf{88.18} / \textbf{86.12}
& \textbf{2.06}
& \textbf{78.17} / \textbf{75.13} &\textbf{3.04} & \textbf{64.50} / \textbf{62.54} &1.96  \\
\bottomrule
\end{tabular*}
\end{table*}

\subsection{Implementation details}\label{implement}
{\color{black}Experiments are conducted under two conditions:
(i)~\textit{Ideal communication}, where transmitted features are assumed to be perfectly received without degradation, and (ii)~\textit{Communication with channel impairments}, where features transmitted by non-ego agents are corrupted by simulated wireless channel effects. The performance gap between these two conditions quantifies the robustness of each cooperative perception model to communication-induced feature degradation.

The channel simulation model described in Section~\ref{PF} follows the IEEE~802.11p standard and captures the fundamental effects of wireless propagation on intermediate feature transmission.
Communication parameters are selected based on typical vehicular configurations reported in the literature and standardization documents~\cite{ref41,ref42,ref43}, and are summarized in Table~\ref{tab:comm_params}.
The carrier frequency is set to $5.9$\,GHz, corresponding to the ITS spectrum used by DSRC and C-V2X systems, with a channel bandwidth of $B = 10$\,MHz following the standard vehicular channel configuration.
The transmit power is $P_{\mathrm{tx}} = 23$\,dBm.
Antenna gains are set to $G_{\mathrm{tx}} = 3$\,dBi for vehicles and $6$\,dBi for RSUs, reflecting representative configurations in the literature.
The receiver noise figure is $\mathrm{NF} = 6$\,dB, and the thermal noise power spectral density is $N_0 = -174$\,dBm/Hz.}

All models use PointPillars~\cite{ref35} as the encoder backbone, with a voxel resolution of $0.4$\,m along both spatial dimensions and a maximum of 32 points per voxel.
Training uses the Adam optimiser with an initial learning rate of $1\times10^{-3}$; loss coefficients are set to $\alpha = 1$, $\beta = 2$, and $\gamma = 10{,}000$.
The batch size is 2 for V2XSet and 4 for DAIR-V2X. Detection ranges are $x \in [-140.8,\,140.8]$\,m, $y \in [-38.4,\,38.4]$\,m for V2XSet, and $x \in [-100.8,\,100.8]$\,m, $y \in [-40,\,40]$\,m for DAIR-V2X.
{\color{black}The number of collaborating agents in our experiments follows the standard benchmark settings. Specifically, DAIR-V2X considers vehicle–infrastructure cooperation with two agents, while V2XSet includes up to five agents per scene, consistent with the original protocol \cite{ref10}. This configuration reflects a typical number of effective collaborators within practical communication ranges.

Training is performed on a High-Performance Computing (HPC) cluster using nodes equipped with NVIDIA H100 GPUs (80\,GB) and Intel Xeon Platinum 8468 CPUs.
Inference and validation are conducted on a local workstation with two NVIDIA RTX~5000 Ada GPUs (32\,GB) and an Intel Xeon W-2445 CPU @ 2.10\,GHz.}

\subsection{Evaluation Metrics and Baselines}

Detection performance is evaluated using Average Precision (AP) at IoU thresholds of 0.50 and 0.70.
\textcolor{black}{We compare AFFormer against two categories of representative methods. The first comprises state-of-the-art cooperative perception methods that do not explicitly address communication corruption: V2X-ViT~\cite{ref10}, CoAlign~\cite{ref16}, and MKD-Cooper~\cite{ref20}.
The second comprises corruption-aware methods that explicitly handle degraded features: DSRC~\cite{ref21} and V2VAM+LCRN~\cite{ref31}, which incorporate mechanisms to mitigate or reconstruct corrupted features under degraded conditions.}

To ensure a fair robustness evaluation, all baseline methods are trained and tested under identical communication-corruption conditions, in which the simulated wireless channel degrades transmitted intermediate features before they reach the fusion module.
This protocol enables a consistent assessment of how each framework behaves under feature corruption, regardless of whether the original method was designed to handle communication impairments.
Results for V2X-ViT and V2VAM+LCRN are reproduced using the OpenCOOD framework~\cite{ref44}; results for all other baselines are obtained from their official implementations.

\begin{figure*}[th]
\centering
\includegraphics[width=\textwidth]{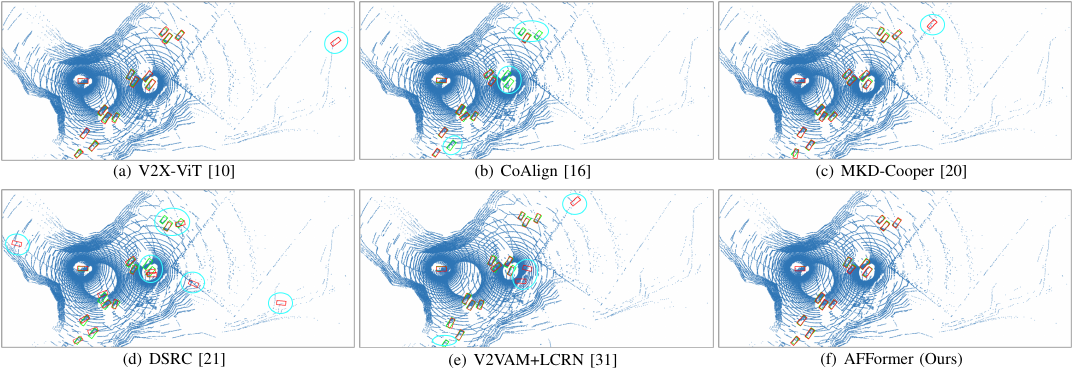} 
\caption{\textcolor{black}{Visualization of detection results on V2XSet under communication-impaired conditions.
Red and green boxes denote detection results and ground-truth, respectively. Some representative false-detection examples are highlighted with bright blue oval circles.}}
\label{fig:vis_v2xset}
\end{figure*}

\begin{figure*}[th]
\centering
\includegraphics[width=\textwidth]{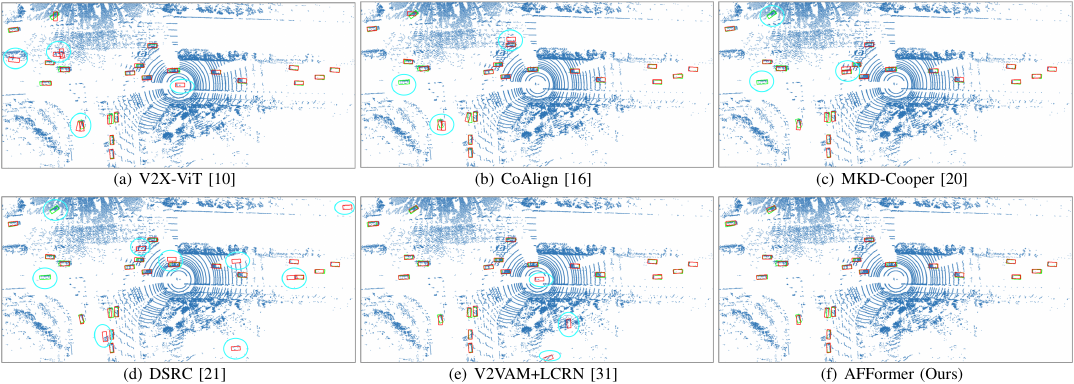} 
\caption{\textcolor{black}{Visualization of detection results on DAIR-V2X under communication-impaired conditions.
Red and green boxes denote detection results and ground-truth, respectively. Some representative false-detection examples are highlighted with bright blue oval circles.}}
\label{fig:vis_dairv2x}
\end{figure*}

\subsection{Results and Discussions}\label{results}
{\color{black}
\subsubsection{Detection Performance}
Table~\ref{tab:performance} reports the AP results at IoU thresholds of 0.5 and 0.7 under both ideal and communication-impaired transmission conditions on the V2XSet and DAIR-V2X datasets.
The performance drop ($\Delta$) is used to quantify robustness to communication-induced feature degradation, and methods marked as corruption-aware are explicitly designed to address such degradation.
The best-performing results are highlighted for reference.

AFFormer consistently achieves the highest detection performance across both datasets under both ideal and impaired communication conditions.
This improvement is mainly attributed to its enhanced multi-agent feature fusion design, which enables more effective interactions across the agent, temporal, and spatial dimensions, thereby improving the quality of the aggregated representations.
Compared with existing methods that rely on simpler aggregation strategies or local feature refinement, AFFormer is better able to exploit complementary information across agents, leading to improved perception performance even under ideal communication conditions.
Specifically, AFFormer attains a peak AP@0.5 of 96.44 on V2XSet and 78.17 on DAIR-V2X under perfect transmission, outperforming the competing methods on both datasets.
In terms of robustness, AFFormer demonstrates strong resilience on V2XSet, limiting the performance drop to only 3.10 and 2.06 for AP@0.5 and AP@0.7, respectively.
Although V2VAM+LCRN achieves a marginally smaller degradation rate on DAIR-V2X at AP@0.7, AFFormer maintains a higher absolute accuracy.
Taken together, these results indicate that AFFormer improves both perception accuracy and robustness to communication-induced corruption.

To further evaluate the effectiveness of AFFormer under communication-impaired conditions, Figs.~\ref{fig:vis_v2xset} and~\ref{fig:vis_dairv2x} present qualitative visualizations on V2XSet and DAIR-V2X, respectively.
On V2XSet, V2X-ViT and MKD-Cooper produce notable false positives in distant scene regions, indicating limited robustness to feature degradation.
CoAlign fails to suppress noise effectively, resulting in missed detections of several distant and occluded objects.
Corruption-aware methods such as DSRC and V2VAM+LCRN exhibit localization instability, manifested by phantom detections in empty regions and redundant misaligned bounding boxes.
These artifacts become more pronounced in the real-world DAIR-V2X scenarios, where each baseline produces at least three false detections; even corruption-aware methods struggle with the increased scene complexity, generating false positives in roadside vegetation and open-road areas.
In contrast, AFFormer produces detections that remain highly consistent with the ground truth across both datasets, successfully suppressing the false-detection artifacts observed in competing methods and confirming its effectiveness in preserving critical features under challenging V2X communication conditions.

\begin{figure*}[!t]
\centering
\includegraphics[width=\textwidth]{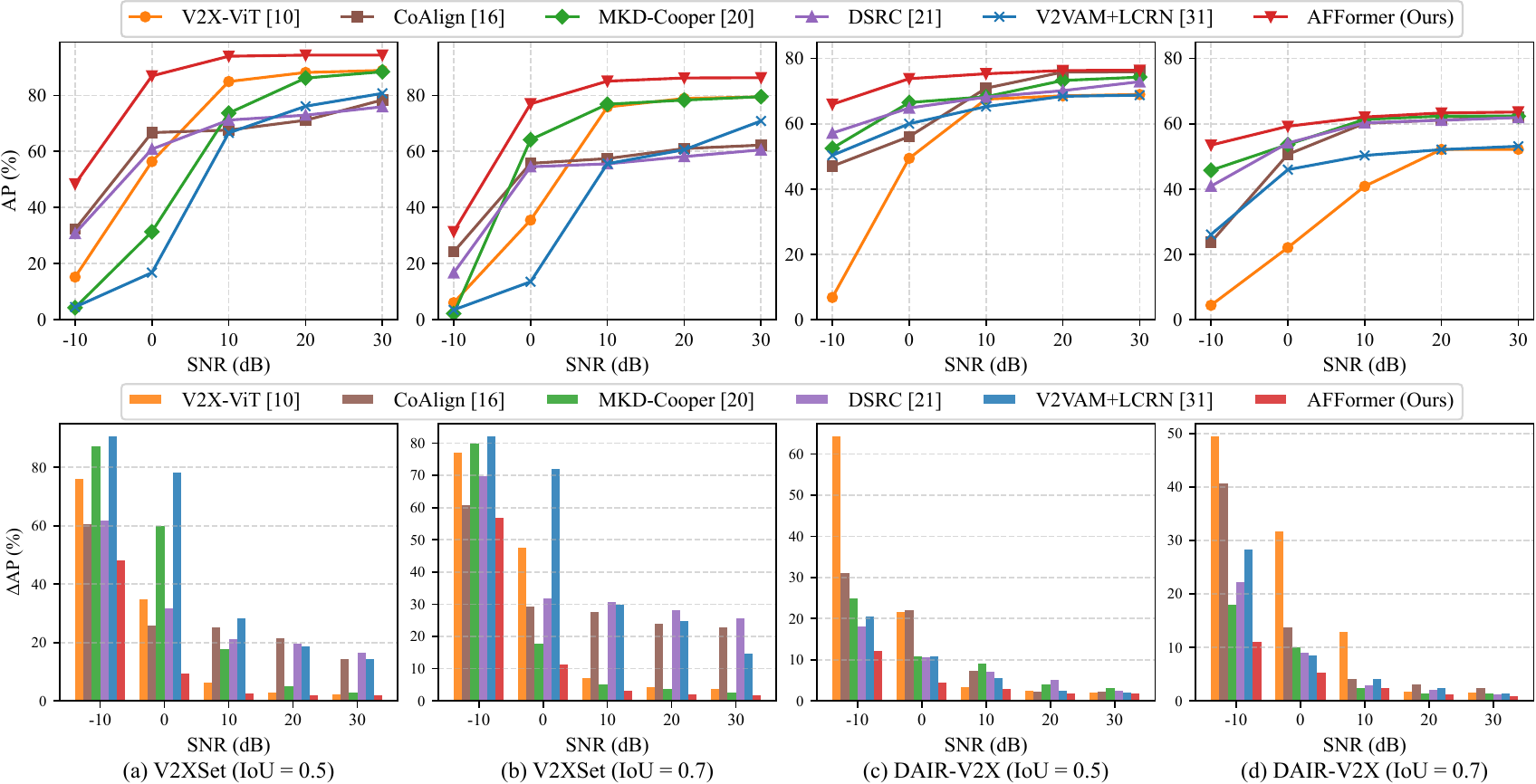}
\caption{\textcolor{black}{Detection performance and AP reduction relative to ideal communication across varying SNR levels on both datasets. The performance reduction is defined as $\Delta \mathrm{AP} = \mathrm{AP}_\mathrm{ideal} - \mathrm{AP}_\mathrm{SNR}$.}}
\label{fig:snr_effect}
\end{figure*}

\begin{figure*}[!t]
\centering
\includegraphics[width=\textwidth]{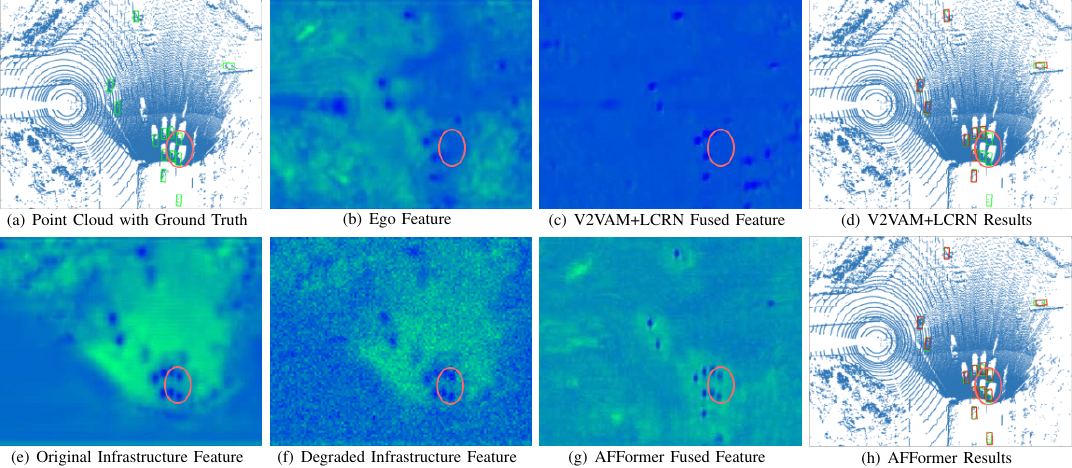}
\caption{\textcolor{black}{Qualitative comparison of fused features and detection outputs on 
DAIR-V2X under channel impairments. 
(a) Input point cloud with ground-truth annotations. 
(b) Ego vehicle feature map. 
(c)--(d) Fused feature map and detection results of V2VAM+LCRN. 
(e) Original infrastructure feature prior to transmission. 
(f) Degraded infrastructure feature received by the ego vehicle under 
channel impairments. 
(g)--(h) Fused feature map and detection results of AFFormer. 
The red ellipse highlights the region of interest containing objects 
occluded from the ego vehicle's perspective. 
Predicted bounding boxes are shown in red; ground-truth boxes are shown 
in green.}}
\label{fig:feature}
\end{figure*}

\subsubsection{Performance under Varying SNRs}
To evaluate robustness under varying channel conditions, the computed SNR in Equation~(\ref{eq:snr_db}) is replaced with fixed values ranging from $-10$\,dB to $30$\,dB in increments of $10$\,dB.
The first row of Fig.~\ref{fig:snr_effect} presents the detection performance of each method across all SNR levels.
To further quantify sensitivity to feature degradation, the AP drop relative to the ideal communication setting is computed as $\Delta \mathrm{AP} = \mathrm{AP}_\mathrm{ideal} - \mathrm{AP}_\mathrm{SNR}$, where $\mathrm{SNR} \in \{-10, 0, 10, 20, 30\}$\,dB.
The corresponding results are shown in the second row of Fig.~\ref{fig:snr_effect}.

As expected, all methods exhibit monotonically improved performance as the SNR increases, confirming the adverse effect of channel noise on the quality of cooperative features.
AFFormer consistently achieves the highest AP across all SNR regimes on both datasets.
This advantage is particularly pronounced under severely impaired conditions at $-10$\,dB, where AFFormer maintains a substantial margin over all baselines, while competing methods, especially V2X-ViT, suffer severe degradation.
As the SNR increases, the performance gap gradually narrows; however, AFFormer retains a consistent lead throughout.
This trend demonstrates the capability of AFFormer to suppress noise-induced feature distortion adaptively and maintain stable cooperative perception across the full SNR range.

The $\Delta \mathrm{AP}$ histograms in the second row of Fig.~\ref{fig:snr_effect} further support these observations.
AFFormer yields the smallest $\Delta \mathrm{AP}$ at every SNR level on both datasets, indicating the lowest sensitivity to feature corruption.
By contrast, the baseline methods exhibit substantially larger performance drops, particularly at low SNR, where the degradation exceeds $60\%$ in several cases.
Moreover, although the $\Delta \mathrm{AP}$ of the baseline methods decreases gradually as the SNR increases, AFFormer maintains consistently low degradation across all regimes, indicating a more stable and noise-resilient feature representation.
This behavior is observed consistently across both datasets and IoU thresholds, further confirming that AFFormer effectively suppresses the propagation of corrupted features and enhances perceptual reliability under unreliable V2X communication conditions.

\subsubsection{Feature Fusion Quality}
Beyond detection accuracy, it is also important to assess the quality of feature fusion under lossy communication conditions.
To this end, the fused feature maps and detection outputs of V2VAM+LCRN and AFFormer on DAIR-V2X under channel impairments are qualitatively compared in Fig.~\ref{fig:feature}.

As shown in Figs.~\ref{fig:feature}(b) and~\ref{fig:feature}(e), the ego vehicle's feature map does not capture the objects located within the red-ellipse region, whereas the original infrastructure feature contains clear object-related responses in that area.
Although the infrastructure feature is degraded during transmission, the received feature in Fig.~\ref{fig:feature}(f) still preserves residual object information that can potentially be exploited during fusion.

The fused feature maps in Figs.~\ref{fig:feature}(c) and~\ref{fig:feature}(g) reveal a clear qualitative difference between the two methods.
V2VAM+LCRN fails to recover the object signatures preserved in the degraded infrastructure feature, producing a fused representation that remains largely similar to the ego-only feature and consequently missing the occluded objects in its detection output, as shown in Fig.~\ref{fig:feature}(d).
By contrast, AFFormer produces a denser and semantically richer fused feature map, with clearly activated responses in the previously occluded region and stronger foreground--background contrast throughout.
This improvement is directly reflected in the detection output in Fig.~\ref{fig:feature}(h), where the occluded objects are successfully localized and spurious detections are suppressed.
These results confirm that AFFormer integrates corrupted infrastructure features more effectively during fusion, thereby enabling more reliable perception under challenging lossy V2X communication conditions.

\begin{table}[!t]
\centering
\tabcolsep=6pt
\renewcommand\arraystretch{1.2}
\caption{\textcolor{black}{Ablation study results under communication impairments.}}
\label{tab:ablation}
\begin{tabular*}{\columnwidth}{@{\extracolsep{\fill}}ccccc}
\toprule
\multirow{2}{*}{Method}&\multicolumn{2}{c}{V2XSet}&\multicolumn{2}{c}{DAIR-V2X} \\
\cmidrule(l{1pt}r{2pt}){2-3} \cmidrule(l{2pt}r{1pt}){4-5}
&AP@0.5$\uparrow$ &AP@0.7$\uparrow$ &AP@0.5$\uparrow$ &AP@0.7$\uparrow$ \\
\midrule
\textit{No fusion}            &81.00 &65.92 &67.91 &49.67 \\
\textit{No historical clues}  &92.92 &84.43 &74.54 &61.51 \\
AFFormer$\setminus$MATA       &41.24 &22.34 &56.22 &34.31 \\
AFFormer$\setminus$DualSA     &93.27 &85.95 &74.24 &59.76 \\
AFFormer$\setminus$UGF        &92.88 &84.97 &74.87 &61.31 \\
AFFormer$\setminus$KD         &92.63 &83.78 &74.37 &55.65 \\
AFFormer                      &\textbf{93.34} &\textbf{86.12} &\textbf{75.13} &\textbf{62.54} \\
\bottomrule
\end{tabular*}
\end{table}
\begin{table}[!t]
\centering
\tabcolsep=6pt
\renewcommand\arraystretch{1.2}
\caption{\textcolor{black}{Computational cost and detection performance of DualSA versus traditional self-attention, evaluated on DAIR-V2X when embedded in AFFormer.}}
\label{tab:dualsa}
\begin{tabular*}{\columnwidth}{@{\extracolsep{\fill}}cccccc}
\toprule
\multirow{2.5}{*}{Module} &\multirow{2.5}{*}{\makecell[c]{FLOPs\\(G)}} 
& \multirow{2.5}{*}{\makecell[c]{Latency\\(ms)}} 
& \multicolumn{2}{c}{Embedded in AFFormer} \\
\cmidrule(l{1pt}r{1pt}){4-5}
& & &AP@0.5$\uparrow$ &AP@0.7$\uparrow$  \\
\midrule
Traditional SA  &43.45 &32.56 &74.24 &59.76 \\
DualSA          &\textbf{14.78} &\textbf{3.48} &\textbf{75.13} &\textbf{62.54} \\
\bottomrule
\end{tabular*}
\end{table}

\subsubsection{Ablation Study}
To evaluate the contribution of each component in AFFormer, six ablation variants are designed by removing or replacing a single module while keeping all other components unchanged.
\textit{No fusion} removes the inter-agent attention (IAT) from the MATA module and relies solely on the ego feature sequence.
\textit{No historical clues} excludes the temporal attention (TAT), using only current-timestep features for fusion.
AFFormer$\setminus$MATA replaces the entire MATA module with max pooling across agents and time.
AFFormer$\setminus$DualSA substitutes the DualSA module with standard self-attention.
AFFormer$\setminus$UGF removes the UGF module and replaces it with element-wise addition for combining spatial-aware features.
Finally, AFFormer$\setminus$KD is trained without knowledge distillation, thereby isolating the contribution of the teacher--student training paradigm.

Table~\ref{tab:ablation} reports the performance of all variants on both datasets under communication impairments.
The \textit{No fusion} baseline performs poorly on both datasets, achieving 81.00/65.92 on V2XSet and 67.91/49.67 on DAIR-V2X, confirming that ego-only perception is insufficient, particularly on the more challenging real-world DAIR-V2X benchmark.
Introducing multi-agent fusion without temporal modeling (\textit{No historical clues}) yields substantial gains, demonstrating the importance of inter-agent feature aggregation.
The remaining gap relative to the full model indicates that historical context provides complementary information not captured by single-timestep fusion alone.

AFFormer$\setminus$MATA suffers the most severe degradation among all variants, dropping to 41.24/22.34 on V2XSet, which is significantly lower than all other configurations.
This collapse indicates that the MATA module is indispensable for jointly modeling inter-agent and temporal correlations.
AFFormer$\setminus$DualSA incurs only minor degradation on V2XSet but more pronounced declines on DAIR-V2X, particularly at AP@0.7 (from 62.54 to 59.76), suggesting that the decomposed spatial attention in DualSA is especially beneficial for handling the greater scene complexity of real-world data.
AFFormer$\setminus$UGF exhibits consistent moderate drops across all settings, demonstrating that element-wise addition is insufficient for filtering unreliable features and that the gating mechanism in UGF provides a necessary adaptive selection of informative signals.
AFFormer$\setminus$KD exhibits a notable decline on DAIR-V2X at AP@0.7 (from 62.54 to 55.65), indicating that knowledge distillation plays an important role in transferring clean-feature representations to the corruption-aware model, particularly under the distribution shift introduced by real-world channel conditions.

To quantitatively evaluate the efficiency of DualSA, it is compared with conventional global self-attention in terms of FLOPs and inference latency under the same experimental setting.
Inference latency is measured with a batch size of 1 and a feature map of size $256\times48\times176$, using 50 warm-up iterations followed by 200 timed runs, with the reported results averaged over all runs.
As shown in Table~\ref{tab:dualsa}, DualSA reduces FLOPs from 43.45\,G to 14.78\,G (approximately 66\% reduction) and latency from 32.56\,ms to 3.48\,ms (approximately 89\% reduction), while simultaneously improving detection performance on DAIR-V2X.
This efficiency gain is attributed to the decomposed attention design of DualSA.
Instead of computing global attention over the full spatial sequence of length $N = H \times W$, DualSA applies attention separately along the horizontal and vertical dimensions, reducing the computational complexity from $\mathcal{O}(H^2W^2)$ to $\mathcal{O}(HW(H+W))$.
These results further demonstrate that DualSA achieves a favorable accuracy-efficiency trade-off, making it well-suited to latency-sensitive cooperative perception applications.

Overall, the complete AFFormer achieves the best performance across all metrics and datasets, demonstrating that each component contributes uniquely to robustness and accuracy, and that their combination provides an effective defense against communication impairments in V2X CP.

\begin{table}[!t]
\centering
\tabcolsep=1pt
\renewcommand\arraystretch{1.2}
\caption{\textcolor{black}{Computational complexity comparison across cooperative perception methods.}}
\label{tab:complexity}
\begin{tabular*}{\columnwidth}{@{\extracolsep{\fill}}cccccc}
\toprule
Method     & \makecell[c]{Params\\(M)} & \makecell[c]{FLOPs\\(G)} 
           & \makecell[c]{Latency\\(ms)} & \makecell[c]{GPU\\Memory (GB)} 
           & \makecell[c]{Inference\\Runtime (ms)} \\
\midrule
V2X-ViT    & 17.31 & 349.54 & 112.52 & 1.66 & 161.54 \\
CoAlign    & 12.90 & 110.77 & \textbf{40.47}  & 0.57 & \textbf{80.42}  \\
MKD-Cooper & \textbf{8.20}  & \textbf{87.89}  & 45.22  & \textbf{0.56} & 86.27 \\
DSRC       & 26.04 & 292.53 & 92.71  & 1.05 & 139.70 \\
V2VAM+LCRN & 18.27 & 303.21 & 103.75 & 2.37 & 170.80 \\
AFFormer   & 16.70 & 492.87 & 78.35  & 1.05 & 113.32 \\
\bottomrule
\end{tabular*}
\end{table}

\subsubsection{Computational Complexity}
Table~\ref{tab:complexity} compares the computational complexity of AFFormer with all baseline methods across five metrics: parameter count, FLOPs, inference latency, GPU memory consumption, and inference runtime.
Inference latency is measured with a batch size of 1 to simulate single-scene real-time inference, following 50 warm-up iterations to stabilize the CUDA runtime and cuDNN autotuning.
The reported latency is the mean over 200 timed iterations using CUDA Event synchronization.
Inference runtime denotes the average per-sample processing time measured over the full test set of DAIR-V2X.

AFFormer incurs the highest FLOPs among all compared methods, reaching 492.87\,G, which is attributable to its multi-dimensional attention mechanism operating jointly across the agent, temporal, and spatial dimensions.
Nevertheless, this theoretical cost does not translate proportionally into wall-clock runtime.
AFFormer achieves an inference latency of 78.35\,ms and an average inference runtime of 113.32\,ms, both of which are lower than those of V2X-ViT and V2VAM+LCRN.
This discrepancy between FLOPs and measured latency can be explained by the decomposed attention design of DualSA, which replaces global attention with complexity $\mathcal{O}(H^2W^2)$ by two sequential passes with complexities $\mathcal{O}(HW^2)$ and $\mathcal{O}(WH^2)$, yielding an overall complexity of $\mathcal{O}(HW(H+W))$ and enabling efficient parallelization on modern GPU hardware.

In terms of parameter count, AFFormer requires 16.70\,M parameters, comparable to V2X-ViT (17.31\,M) and V2VAM+LCRN (18.27\,M), and substantially fewer than DSRC (26.04\,M).
Its GPU memory consumption is 1.05\,GB, identical to that of DSRC and still within a practically manageable range despite the higher FLOPs.
The most lightweight methods, CoAlign and MKD-Cooper, achieve lower FLOPs and memory usage through simpler fusion strategies.
However, as shown in Table~\ref{tab:performance}, this reduction in computational cost is accompanied by lower detection accuracy and weaker robustness under communication impairments.

Overall, while AFFormer is not the most computationally lightweight method, it achieves a competitive runtime profile relative to other corruption-aware and attention-based baselines while delivering superior detection accuracy and robustness.

\section{Conclusion} \label{conclusion}
This paper proposes AFFormer, a Transformer-based feature fusion framework for robust cooperative perception under V2X communication impairments. By jointly modeling inter-agent, temporal, and spatial correlations through the MATA, DualSA, and UGF modules, AFFormer effectively mitigates the impact of corrupted features. Extensive experiments on both V2XSet and DAIR-V2X demonstrate that AFFormer consistently outperforms state-of-the-art methods, validating its effectiveness in improving detection accuracy and robustness under challenging communication conditions.

Despite these promising results, several limitations remain, which warrant further investigation.
The current framework does not explicitly account for communication delays or packet loss, which are common in real-world V2X systems. 
Future work will incorporate delay-aware and packet-loss-resilient mechanisms to enhance robustness under dynamic network conditions.
In addition, current evaluations are primarily constrained by the scale of existing datasets. Extending AFFormer to dense urban scenarios or large platoons may introduce communication bottlenecks. 
Thus, exploring scalable strategies such as agent selection or hierarchical fusion remains a key direction for future work.
Furthermore, AFFormer is evaluated on LiDAR-based BEV feature representations using public cooperative perception benchmarks for object detection. Although the proposed feature-level fusion framework is potentially extensible to other sensing modalities, such as cameras and radar, such heterogeneous settings are not explicitly investigated in this study. Incorporating additional modalities is expected to further enhance perception performance and robustness in complex environments.}

\end{document}